\PassOptionsToPackage{numbers,sort}{natbib}
\PassOptionsToPackage{hyphens}{url}
\documentclass[a4paper,fleqn]{cas-dc}

\usepackage{amsmath}
\usepackage{amssymb}
\usepackage{graphicx}
\usepackage{booktabs}
\usepackage{tabularx}
\usepackage{url}
\usepackage{natbib}
\usepackage{xcolor}
\usepackage{hyperref}
\usepackage{orcidlink}
\hypersetup{hypertexnames=false}
\AtBeginDocument{\hypersetup{colorlinks=true,linkcolor=black,citecolor=black,
  urlcolor=black,pdfborder={0 0 0}}}
\ExplSyntaxOn
\RenewDocumentCommand \firstname {} { \textcolor{black}{ \seq_use:Nn \l_stm_au_seq { ~ } } }
\RenewDocumentCommand \surname {}
  { \l_stm_au_sn_seq
    \tl_if_empty:NF \l_stm_au_orcid_tl { \, \exp_args:NV \orcidlink \l_stm_au_orcid_tl } }
\bool_gset_true:N \g_stm_nologo_bool
\RenewDocumentCommand \orcidauthor { m m }
  { \seq_gput_right:Nn \g_stm_orcid_seq
      { \href{https://orcid.org/#1}{ \ttfamily #1 } \parsename{#2} \space(\eadauthor) } }
\ExplSyntaxOff

\newcommand{\pearsoncorr}{\operatorname{corr}}
\newcommand{\vectorize}{\operatorname{vec}}
\newcommand{\vc}[2]{\begin{tabular}[c]{@{}l@{}}#1\\{\footnotesize[#2]}\end{tabular}}
\newcommand{\tablenote}[1]{\par\smallskip{\footnotesize\raggedright #1\par}}

\shorttitle{Cross-Modal Prediction of fNIRS Haemodynamics From Infra-Slow EEG}
\shortauthors{Jany et al.}

\begin{document}

\title[mode=title]{Cross-Modal Prediction of fNIRS Haemodynamics From Infra-Slow EEG: Ridge Regression, Shared Low-Rank Space and Two-Run Calibration}\tnotemark[1]
\tnotetext[1]{Preprint. This manuscript has been submitted to Computers in Biology and Medicine and has not yet been peer reviewed.}

\author[1,2]{Rafsan Jany}[orcid=0000-0001-8346-6938]
\cormark[1]
\ead{rafsanjany@ust.ac.kr}

\author[3]{Muntasir Hossain}[orcid=0009-0004-8479-8343]
\ead{muntasir21@iut-dhaka.edu}

\author[4]{Maisha Rahman}[orcid=0009-0004-6672-1940]
\ead{maisharahman@iut-dhaka.edu}

\affiliation[1]{organization={University of Science and Technology (UST)},
            addressline={Gajeong-ro, Yuseong-gu},
            city={Daejeon},
            postcode={34113},
            country={Republic of Korea}}
\affiliation[2]{organization={Digital Health Research Division, Korea Institute of Oriental Medicine},
            addressline={1672 Yuseong-daero, Yuseong-gu},
            city={Daejeon},
            postcode={34054},
            country={Republic of Korea}}
\affiliation[3]{organization={Department of Computer Science and Engineering, Islamic University of Technology},
            addressline={Board Bazar},
            city={Gazipur},
            postcode={1704},
            country={Bangladesh}}
\affiliation[4]{organization={Department of Electrical and Electronic Engineering, Islamic University of Technology},
            addressline={Board Bazar},
            city={Gazipur},
            postcode={1704},
            country={Bangladesh}}

\cortext[1]{Corresponding author.}

\begin{abstract}
Objective: Estimating haemodynamic signals from scalp electroencephalography (EEG) would allow low-cost monitoring, but cross-person prediction is close to chance after conventional filtering. We tested whether the band removed by that filtering carries the missing information.
Methods: We analysed simultaneous EEG and functional near-infrared spectroscopy (fNIRS) from 88 people in four cohorts from three laboratories, and EEG with functional magnetic resonance imaging (fMRI) from 27 people. Ridge regression on EEG below 0.5~Hz was compared with band-power models under leave-one-person-out, leave-one-run-out and frozen-transfer validation. Pass rules were set before each test, amended tests are labelled, and Bonferroni correction covered all attempts.
Results: In hBCI, a 0.5--45~Hz band-pass cut cross-person oxyhaemoglobin prediction by a deep network from Pearson $r=0.158$ to $0.019$. Infra-slow EEG gave $r=0.135$--$0.203$ in three cohorts, against $0.006$--$0.045$ for band power; one had poor optical coupling, the advantage was not established in a fourth cohort, and it depended on the analysis form in two. One shared EEG space predicted fNIRS and fMRI in unseen people (fMRI $p=0.0001$) without loss for either. After two calibration runs, EEG-only models exceeded each person's mean response by $0.059$--$0.176$, and the calibration score predicted later accuracy (Spearman $\rho=0.56$--$0.90$). Band power led local oxyhaemoglobin by 6~s in hBCI (29 of 29 people, Wilcoxon $p<0.0001$). Models frozen in one laboratory did not transfer to another.
Conclusions: Infra-slow EEG carries person-specific information about haemodynamics. Calibration with paired recordings is a promising route to EEG-only monitoring.
\end{abstract}


\maketitle

\section{Introduction}
\label{sec:intro}

Scalp electroencephalography (EEG) is inexpensive and portable but does not measure blood oxygenation; functional near-infrared spectroscopy (fNIRS) and functional magnetic resonance imaging (fMRI) do, at higher cost or with less portability \cite{Scholkmann2014,Hillman2014}. The signals are linked through neurovascular coupling \cite{Iadecola2017}. Hybrid EEG--fNIRS has mostly been used to improve brain--computer interface decoding \cite{Ahn2017}. A different goal is to estimate the haemodynamic signal from EEG alone, so that a person could be monitored without optical or magnetic resonance hardware.

Band-limited EEG power correlates with the blood-oxygen-level-dependent (BOLD) signal, negatively for alpha \cite{Goldman2002}, and spontaneous EEG power correlates only weakly with oxyhaemoglobin \cite{Keles2016}. Regional BOLD has been estimated from an EEG fingerprint \cite{MeirHasson2014}, convolutional networks have transcoded between EEG and fMRI \cite{LiuSajda2019}, and a transformer has synthesised resting-state fMRI from EEG in unseen people \cite{Li2024}. For fNIRS, a multimodal autoencoder predicted resting signals from 0.1--100~Hz EEG within each of 40 patients with epilepsy \cite{Sirpal2021}.

EEG is routinely high-pass filtered at 0.1--1~Hz, which attenuates infra-slow activity, and filter settings can alter conclusions \cite{Widmann2015}. Infra-slow fluctuations relate to cortical excitability \cite{Vanhatalo2004} and fMRI dynamics \cite{Hiltunen2014}, and EEG oscillations near 0.1~Hz are coherent with NIRS signals \cite{Nikulin2014}. Their origin is debated: scalp direct-current shifts can have non-neuronal generators \cite{Voipio2003} and relate to respiration \cite{Yuan2013}, while systemic physiology contaminates fNIRS \cite{Tachtsidis2016}. Global and local signals must therefore be separated.

To our knowledge, no study has tested whether the band removed by conventional filtering carries cross-person information about fNIRS, compared infra-slow EEG with band power across cohorts and laboratories, asked whether a short calibration predicts how well EEG-only monitoring will work for a given person, or used one EEG representation to predict both fNIRS and fMRI. Earlier work is within person \cite{Sirpal2021}, targets a single modality \cite{MeirHasson2014,LiuSajda2019,Li2024}, or describes coupling without prediction \cite{Keles2016,Nikulin2014,Moore2021}.

We address these gaps with a band-resolved evaluation of EEG-to-oxyhaemoglobin prediction in four EEG--fNIRS cohorts from three laboratories; a shared EEG subspace for fNIRS and fMRI, tested for non-inferiority; two-run calibration with a person-level reliability score; tests of the band-power coupling kernel; and a dated analysis plan with correction across all attempts.

\section{Materials and methods}
\label{sec:methods}

\begin{figure*}[!t]
\centering
\includegraphics[width=\linewidth]{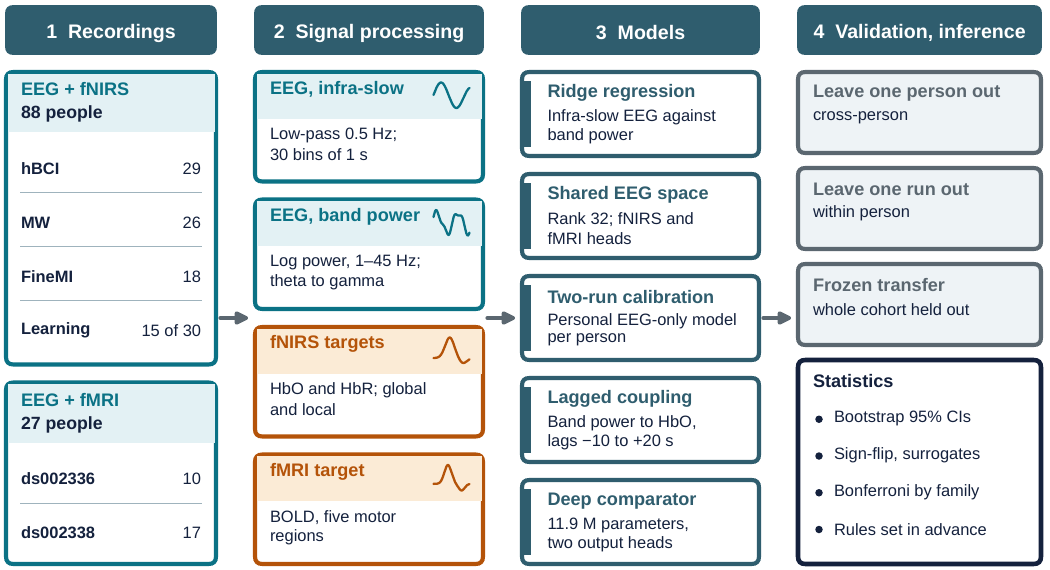}
\caption{Analysis pipeline. Small traces are schematic, not data. HbO, HbR: oxy- and deoxyhaemoglobin; BOLD: blood-oxygen-level-dependent signal; CI: confidence interval.}
\label{fig:pipeline}
\end{figure*}

Figure~\ref{fig:pipeline} summarises the analysis.

\subsection{Datasets}
\label{sec:data}

All data are public (Table~\ref{tab:data}): 88 people with EEG--fNIRS, 73 in the main analyses and 15 in a confirmation cohort from a third laboratory, and 27 with EEG--fMRI. hBCI and MW come from TU Berlin and may share participants (Section~\ref{sec:transfer}). The FineMI descriptor reports a 0.5--100~Hz band-pass \cite{Yi2025}, but the file headers show no high-pass filter and EEG power rose steeply below 0.5~Hz. Fifteen of the 30 learning-capability participants passed synchronisation and length rules set before any EEG--fNIRS relation was computed; an early band-power baseline had been run, so this cohort was not entirely unseen. HEFMI-ICH \cite{Shi2025} and OpenNeuro ds002739 were excluded beforehand because their published EEG has no power below 0.5~Hz. Quality and preprocessing details are in the Supplementary Material (Tables S1--S2).

\begin{table*}[!t]
\centering
\caption{Datasets.}
\label{tab:data}
\small
\begin{tabularx}{\linewidth}{lcXXX}
\toprule
Cohort & $N$ & Task & EEG & Haemodynamics \\
\midrule
hBCI \cite{Shin2017} & 29 & MI and mental arithmetic, 6 runs & 30 ch, BrainAmp, TC 10 s & fNIRS, 36 ch, 10 Hz \\
MW \cite{Shin2018} & 26 & $n$-back, 3 runs & BrainAmp, TC 10 s & fNIRS \\
FineMI \cite{Yi2025} & 18 & Eight joint MI tasks, 8 blocks & 60 scalp ch, Neuroscan, 1000 Hz & fNIRS, NIRScout, 24 ch, 7.8 Hz \\
Learning capability \cite{Chaiyanan2025} & 30 (15 used) & Implicit learning & 32 ch, Nihon Kohden, 1000 Hz & fNIRS, Hitachi, 10 Hz \\
ds002336, ds002338 \cite{Lioi2020} & 10 + 17 & MI neurofeedback & Scanner-corrected derivative, 200 Hz & fMRI, 5 motor ROIs, TR 2 s \\
\bottomrule
\end{tabularx}
\tablenote{hBCI: hybrid brain--computer interface; MW: mental workload; FineMI: multi-joint upper-limb motor imagery; MI: motor imagery; ch: channels; TC: amplifier time constant; ROI: region of interest; TR: repetition time. Learning capability is the ``third laboratory''.}
\end{table*}

\subsection{Signal processing}
\label{sec:signal}

\begin{figure*}[!t]
\centering
\includegraphics[width=\linewidth]{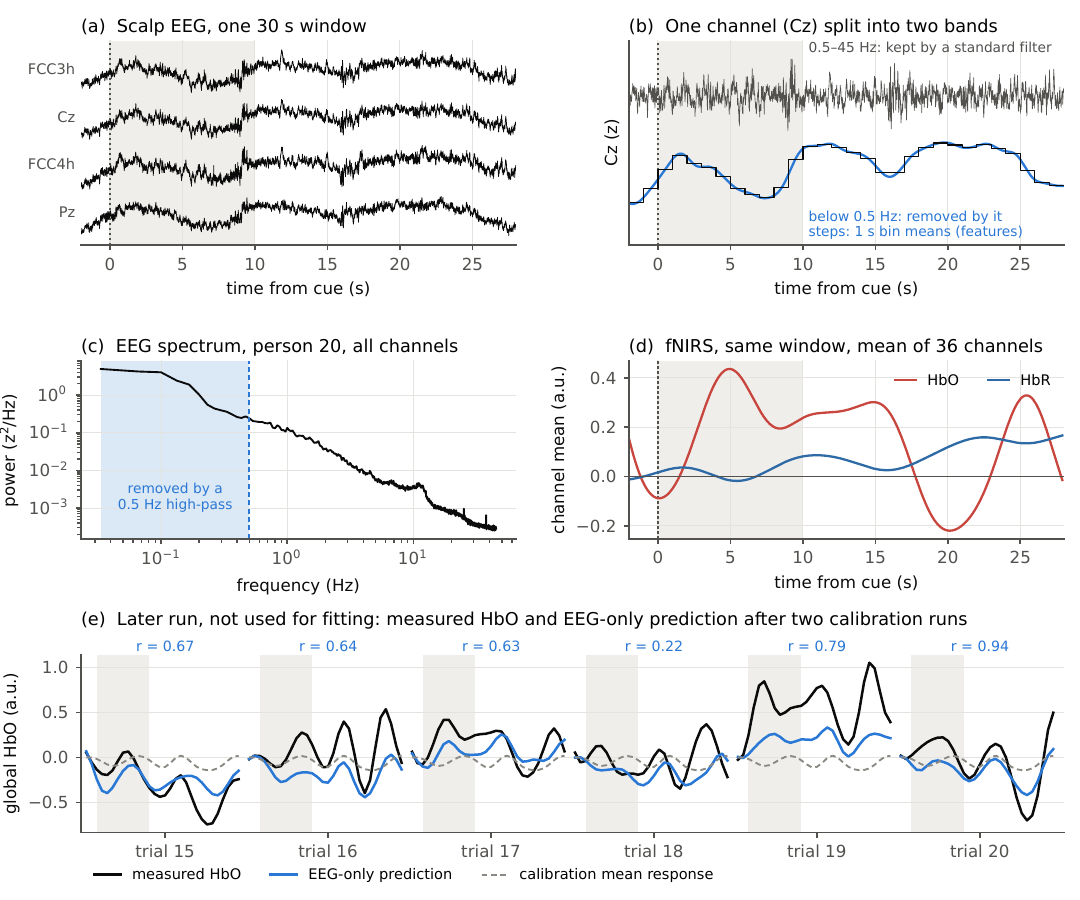}
\caption{Recorded signals from one hBCI person (person 20, whose calibration gain is the median of 29). (a) EEG of one cue-locked window; shading marks the 10~s task. (b) One channel split into the 0.5--45~Hz band and the infra-slow band below 0.5~Hz; steps are the 1~s means used as features. (c) EEG power spectrum over all windows. (d) Channel-mean HbO and HbR of the same window. (e) Channel-mean HbO on a later run against the personal EEG-only prediction of Eq.~(\ref{eq:personal}) calibrated on two earlier runs, and the calibration mean response; $r$ is per trial.}
\label{fig:signals}
\end{figure*}

Figure~\ref{fig:signals} follows one hBCI person through these steps. The raw window (Fig.~\ref{fig:signals}a) mixes fast oscillations with a slow drift of similar size. Splitting a channel at 0.5~Hz (Fig.~\ref{fig:signals}b) separates the two: the part above 0.5~Hz is what a conventional pipeline keeps, and the smooth part below it, summarised by 1~s means, is what our features use. The person's spectrum (Fig.~\ref{fig:signals}c) rises steeply below 0.5~Hz, so most of the recorded EEG power lies in the band that is usually discarded. This person was chosen because their calibration gain was the median of the 29 hBCI participants, not because their prediction was good.

\subsubsection{EEG features}
\label{sec:bandpower}
Trials were cut from $-2$ to $+28$~s around the cue; a stress test used consecutive uncued 30~s windows. EEG was resampled to 200~Hz and z-scored per window. Infra-slow features were EEG low-passed at 0.5~Hz (fourth-order Butterworth, zero phase) and averaged into 30 bins of 1~s per channel, giving $30N_c$ features for $N_c$ channels, z-scored over each person's own windows. Band-power features were log power in five bands from 1 to 45~Hz in 10 bins for the window analysis, and log power in theta, alpha, beta and gamma (4--45~Hz) in 1~s means, band-passed at 0.01--0.1~Hz, for the coupling analysis. Z-scoring each window removes its offset and overall scale, so the infra-slow features describe how the potential changes within the window rather than the absolute electrode offset. In hBCI each person contributed 120 windows from six runs (three of motor imagery and three of mental arithmetic, 20 trials each), 3,480 windows in total.

\subsubsection{Haemodynamic targets}
Oxy- and deoxyhaemoglobin (HbO, HbR) were computed with the modified Beer--Lambert law \cite{Delpy1988} where data were distributed as optical intensities. For the window analysis they were band-passed at 0.01--0.2~Hz, baseline-corrected over the 2~s before the cue and averaged into 30 bins; for the coupling analysis they were band-passed at 0.01--0.1~Hz at 1~Hz. The \emph{global} target is the channel mean and the \emph{local} target is each channel minus that mean; global signal regression was not applied because it removed most of the evoked response. Frozen-transfer and pooled analyses scored pooled channel means of HbO and HbR, so their values are not directly comparable with Table~\ref{tab:slow}. BOLD was averaged over five motor regions, detrended and cut into 15 volumes per trial relative to the first volume; EEG was the distributed scanner-corrected derivative, aligned by volume triggers. Across cohorts, EEG was mapped onto the hBCI montage by spherical-spline interpolation \cite{Perrin1989}, and Euclidean alignment \cite{HeWu2020} was tested for harmonisation.

\subsection{Prediction models}
\label{sec:models}

\begin{figure*}[!t]
\centering
\includegraphics[width=\linewidth]{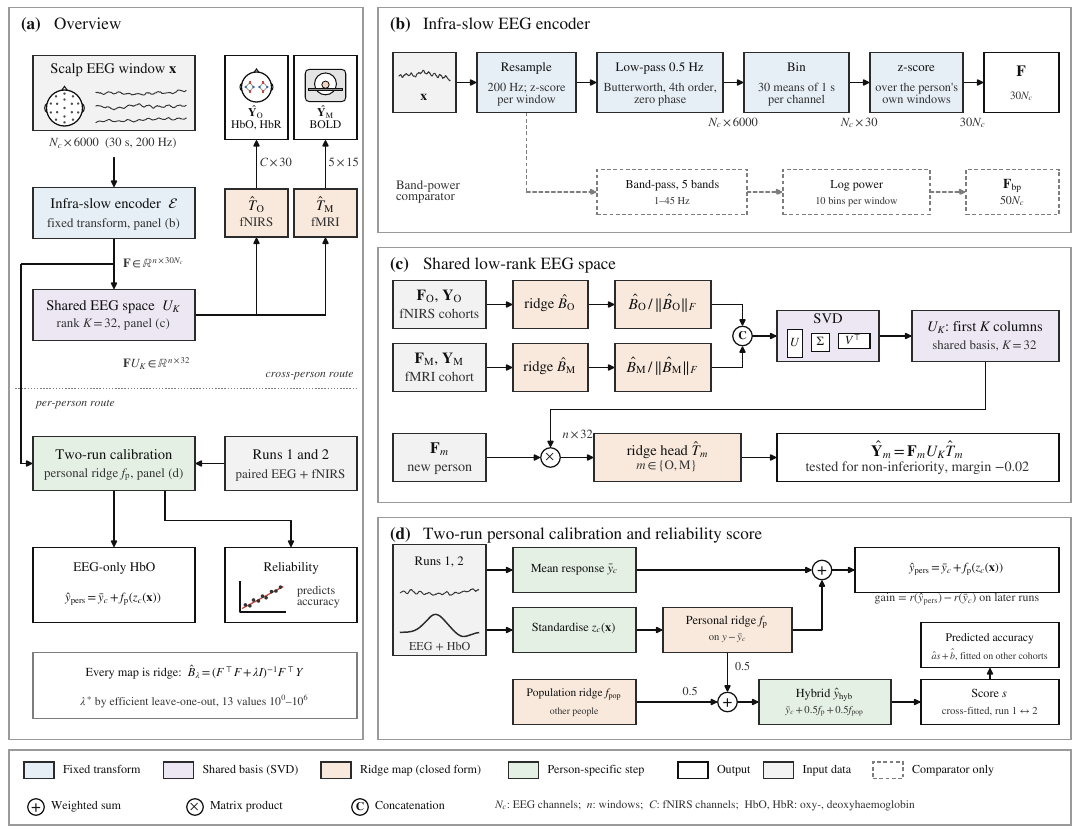}
\caption{Model architecture. (a) Overview: infra-slow encoder, shared EEG space and personal calibration. (b) Infra-slow EEG encoder. (c) Shared low-rank EEG space. (d) Two-run personal calibration and reliability score. Small traces are schematic, not data.}
\label{fig:arch}
\end{figure*}

Figure~\ref{fig:arch} shows the model architecture.

\subsubsection{Ridge regression}
For features $F\in\mathbb{R}^{n\times d}$ and centred targets $Y\in\mathbb{R}^{n\times q}$, ridge regression estimates coefficients by Eq.~(\ref{eq:ridge}):
\begin{equation}
\begin{split}
\hat{B}_{\lambda}&=\arg\min_{B}\ \lVert Y-FB\rVert_F^{2}+\lambda\lVert B\rVert_F^{2}\\
&=(F^{\top}F+\lambda I)^{-1}F^{\top}Y .
\end{split}
\label{eq:ridge}
\end{equation}
One $\lambda$ serves all outputs, chosen by the efficient leave-one-out error of Eq.~(\ref{eq:loo}):
\begin{equation}
\lambda^{\ast}=\arg\min_{\lambda\in\Lambda}\sum_{i=1}^{n}\frac{\lVert y_i-\hat{y}_{i,\lambda}\rVert^{2}}{(1-H_{ii,\lambda})^{2}},
\label{eq:loo}
\end{equation}
where $H_{\lambda}=F(F^{\top}F+\lambda I)^{-1}F^{\top}$ and $\Lambda$ holds 13 log-spaced values from $10^{0}$ to $10^{6}$. Validation left out one person, one run, or a whole cohort with the model frozen (transfer).

\subsubsection{Shared space}
Ridge maps $\hat{B}_{\mathrm{O}}$ (fNIRS) and $\hat{B}_{\mathrm{M}}$ (fMRI) on common EEG features define the shared basis by Eqs.~(\ref{eq:svd})--(\ref{eq:head}):
\begin{equation}
\left[\hat{B}_{\mathrm{O}}/\lVert\hat{B}_{\mathrm{O}}\rVert_F,\ \hat{B}_{\mathrm{M}}/\lVert\hat{B}_{\mathrm{M}}\rVert_F\right]=U\Sigma V^{\top},
\label{eq:svd}
\end{equation}
\begin{equation}
U_K=U_{:,1:K},\qquad \hat{Y}_m=F_m U_K\hat{T}_m,
\label{eq:head}
\end{equation}
where $\hat{T}_m$ is a ridge head fitted on $F_mU_K$ for modality $m\in\{\mathrm{O},\mathrm{M}\}$ and $K=32$. Non-inferiority \cite{Lakens2017} against a modality-specific model required Eq.~(\ref{eq:noninf}):
\begin{equation}
\mathrm{CI}^{90\%}_{\mathrm{low}}\!\left(r_m^{\mathrm{shared}}-r_m^{\mathrm{single}}\right)>-0.02 .
\label{eq:noninf}
\end{equation}
The control permuted fMRI targets within person.

\subsubsection{Personal calibration}
From $k$ calibration runs, with calibration mean response $\bar{y}_{c}$ and features $z_c(x)$ standardised on calibration data, the personal EEG-only model is Eq.~(\ref{eq:personal}):
\begin{equation}
\hat{y}_{\mathrm{pers}}(x)=\bar{y}_{c}+f_{\mathrm{p}}\!\left(z_c(x)\right),
\label{eq:personal}
\end{equation}
where $f_{\mathrm{p}}$ is a ridge model of the residual. The gain is its score minus that of replaying $\bar{y}_{c}$. The reliability analysis used the hybrid of Eq.~(\ref{eq:hybrid}), averaging personal and population ridge models:
\begin{equation}
\begin{split}
\hat{y}_{\mathrm{hyb}}(x)={}&\bar{y}_{c}+0.5\,f_{\mathrm{p}}\!\left(z_c(x)\right)\\
&+0.5\,f_{\mathrm{pop}}\!\left(z_c(x)\right).
\end{split}
\label{eq:hybrid}
\end{equation}
Each run of a person was held out in turn. Calibration runs were drawn from that person's remaining runs, with up to ten random subsets for each number of runs, and the baseline was the task-specific mean response of the calibration runs. The population model was fitted on all other people, with each person's features standardised on their own windows and targets taken relative to their own task mean, so it carried no information about the held-out person. The calibration score was cross-fitted between the two calibration runs, and accuracy came from the person's remaining runs. A least-squares map from calibration score to later accuracy was fitted on the other cohorts and tested on the held-out cohort against a constant prediction.

\subsubsection{Lagged coupling}
Within each person, with run-standardised band power $\mathbf{p}(t)\in\mathbb{R}^{4N_c}$ and local HbO $\mathbf{y}(t)$, Eq.~(\ref{eq:lag}) predicts HbO $\tau$ seconds after the EEG, so a positive lag means that EEG leads:
\begin{equation}
\begin{split}
\hat{\mathbf{y}}(t+\tau)&=W_{\tau}^{\top}\mathbf{p}(t)+\mathbf{b}_{\tau},\\
\tau&\in\{-10,\dots,20\}\,\mathrm{s}.
\end{split}
\label{eq:lag}
\end{equation}
$W_{\tau}$ is fitted by Eq.~(\ref{eq:ridge}) on all runs except run $j$. Eqs.~(\ref{eq:curve})--(\ref{eq:delay}) give the lag curve $R(\tau)$ over $J$ runs and $C$ channels, and the delay:
\begin{equation}
\begin{split}
R(\tau)={}&(1/J)\sum_{j=1}^{J}(1/C)\\
&\times\sum_{c=1}^{C}\pearsoncorr\!\left(\hat{y}^{(-j)}_{c},\,y_{c}\right)_{\tau},
\end{split}
\label{eq:curve}
\end{equation}
\begin{equation}
\hat{\tau}=\arg\max_{\tau}R(\tau),\qquad R_{\max}=\max_{\tau}R(\tau).
\label{eq:delay}
\end{equation}
Surrogates shift HbO circularly by $\delta_k\in[60,T-60]$~s, $k=1,\dots,100$, and a person beats chance when Eq.~(\ref{eq:surr}) holds:
\begin{equation}
p_{\mathrm{null}}=\big(1+\#\{k:R^{(\delta_k)}_{\max}\geq R_{\max}\}\big)/101<0.05 .
\label{eq:surr}
\end{equation}
Kernel reproducibility, Eq.~(\ref{eq:kappa}), compares curves $R_A$ and $R_B$ from the two halves of a recording against $S=20$ shifts of the second half, tested across people with the Wilcoxon signed-rank test:
\begin{equation}
\begin{split}
\kappa={}&\pearsoncorr(R_A,R_B)\\
&-(1/S)\sum_{k=1}^{S}\pearsoncorr\big(R_A,R_B^{(\delta_k)}\big).
\end{split}
\label{eq:kappa}
\end{equation}
In a control, the cue-locked response was removed by keeping the residual $\boldsymbol{\varepsilon}$ of the finite-impulse-response model in Eq.~(\ref{eq:fir}), with cue indicator $u(t)$:
\begin{equation}
\mathbf{y}(t)=\boldsymbol{\beta}_0+\sum_{l=0}^{29}\boldsymbol{\beta}_l\,u(t-l)+\boldsymbol{\varepsilon}(t).
\label{eq:fir}
\end{equation}
Two half-recording delays agreed when they differed by at most 2~s. The lag analysis was repeated with HbR as the target; a later HbR peak (by 1--2~s) was expected.

\subsubsection{Deep comparator}
For EEG $\mathbf{x}$ of dataset $D$ and person $s$, Eqs.~(\ref{eq:enc})--(\ref{eq:out}) give the encoder and output heads:
\begin{equation}
\begin{split}
\mathbf{h}_0&=A_D\mathbf{x},\\
\mathbf{h}_{\ell}&=\mathrm{GELU}\big(\mathrm{BN}(\mathrm{Conv}_{\ell}(\mathbf{h}_{\ell-1}))\big),
\end{split}
\label{eq:enc}
\end{equation}
\begin{equation}
\mathbf{g}=\big[\vectorize(\mathrm{Pool}_{16}(\mathbf{h}_3));\ \mathbf{e}_s\big]\in\mathbb{R}^{1032},
\label{eq:z}
\end{equation}
\begin{equation}
\begin{split}
\mathbf{u}&=\mathrm{Drop}_{0.3}\big(\mathrm{GELU}(W_1\mathbf{g})\big),\\
\boldsymbol{\mu}&=W_{\mu}\mathbf{u},\qquad \mathbf{v}=\mathrm{clip}(W_{v}\mathbf{u},-8,8),
\end{split}
\label{eq:out}
\end{equation}
where $A_D$ maps the montage to 64 channels, $\mathrm{Conv}_{\ell}$ ($\ell=1,2,3$) has 64 filters, kernel 25, 15, 9 and stride 5, 4, 3, $W_1$ has 256 units, $\mathbf{e}_s$ is an 8-value subject embedding (the mean for unseen people), and $\boldsymbol{\mu}$ and $\mathbf{v}$ are the predicted mean and log-variance. The loss for modality $m$ is Eq.~(\ref{eq:loss}), with bars denoting means over channels and time:
\begin{equation}
\begin{split}
\mathcal{L}_m={}&1-\overline{\pearsoncorr(\boldsymbol{\mu},\mathbf{y})}\\
&+0.1\,\overline{\exp(-\mathbf{v})(\boldsymbol{\mu}-\mathbf{y})^{2}+\mathbf{v}}\\
&+0.05\,\overline{(\mu_{t+1}-\mu_{t})^{2}} .
\end{split}
\label{eq:loss}
\end{equation}
Training minimised $\mathcal{L}_{\mathrm{O}}+\mathcal{L}_{\mathrm{M}}$ with AdamW (learning rate $10^{-3}$, batch 64, 30 epochs) over five person-wise folds; the network had 11.9 million parameters. Uncertainty used Monte Carlo dropout \cite{Gal2016}.

\subsection{Evaluation and statistics}
\label{sec:stats}

The window score in Eq.~(\ref{eq:score}) was computed after demeaning each prediction and target within person and task, so a model could not score by replaying a person's average response:
\begin{equation}
\begin{split}
r_{\mathrm{subject}}={}&(1/N)\sum_{s=1}^{N}(1/|\mathcal{W}_s|)\\
&\times\sum_{w\in\mathcal{W}_s}\pearsoncorr\!\left(\vectorize\hat{Y}^{\ast}_{w},\ \vectorize Y^{\ast}_{w}\right),
\end{split}
\label{eq:score}
\end{equation}
where $\mathcal{W}_s$ is the set of windows of person $s$ and the asterisk denotes demeaned values. Calibration analyses keep the task mean, since replaying it is their baseline. The person was the statistical unit, and confidence intervals (CIs) are 95\% bootstrap intervals over people. Significance used a sign-flip test against a within-person pairing null; a prediction was detectable when $p<0.05$, the CI lower bound exceeded the null mean and $r\geq0.05$. Reliability required Spearman $\rho>0.2$ with a CI above zero. For identification, each person's held-out run was assigned to the person whose leave-one-run-out model predicted it best. For decoding, out-of-fold predicted haemodynamics fed a logistic-regression classifier, compared with principal-component and random-projection codes of the same dimension.

\subsection{Analysis plan and multiplicity}
\label{sec:plan}

Pass rules were written into a dated internal plan, not a public preregistration, before each test; later decisions are labelled amendments or post hoc. Bonferroni correction counted every attempt at a claim within its family: 8 for the shared space ($\alpha=0.0063$), 3 within and 16 across laboratories for transfer ($0.017$, $0.0031$), 14 for coupling ($0.0036$), 5 for decoding ($0.010$) and 4 for reliability ($0.0125$). Later rounds formed their own families: third-laboratory confirmation (6 tests, $0.0083$), method check (7, $0.0071$), stress tests (12, $0.0042$) and strengthening attempts (15, $0.0033$). Results passing only at $p<0.05$ are called nominal (ledger: Supplementary Table S7).

\section{Results}
\label{sec:results}

\subsection{The information lies in the infra-slow band}
\label{sec:slow}
\begin{table*}[!t]
\centering
\caption{Cross-person prediction of HbO by infra-slow EEG and by band power.}
\label{tab:slow}
\footnotesize
\setlength{\tabcolsep}{4pt}
\begin{tabularx}{\linewidth}{lcXXXXXl}
\toprule
 & & \multicolumn{2}{c}{Cue-locked windows} & \multicolumn{3}{c}{Uncued windows} & Identification \\
\cmidrule(lr){3-4}\cmidrule(lr){5-7}
Cohort & $N$ & Slow $r$ & Band $r$ & Slow $r$ & Band $r$ & Difference & own model (chance) \\
\midrule
hBCI & 29 & 0.197 & $\approx$0.02 & \vc{0.203}{0.166, 0.241} & 0.070 & \vc{+0.133}{0.099, 0.170} & 93\% (3\%) \\
MW & 26 & \vc{0.135}{0.089, 0.183} & \vc{0.006}{$-0.015$, 0.026} & \vc{0.147}{0.105, 0.193} & 0.063 & \vc{+0.084}{0.043, 0.128} & 62\% (4\%) \\
FineMI & 18 & \vc{0.203}{0.164, 0.248} & \vc{0.045}{0.036, 0.055} & \vc{0.193}{0.153, 0.239} & 0.069 & \vc{+0.124}{0.082, 0.174} & 83\% (6\%) \\
Third laboratory & 15 & not run & not run & \vc{$-0.003$}{$-0.015$, 0.009} & 0.004 & \vc{$-0.007$}{$-0.023$, 0.007} & not run \\
\bottomrule
\end{tabularx}
\tablenote{Slow: ridge on EEG below 0.5~Hz. Band: ridge on band power 1--45~Hz. Uncued: consecutive 30~s windows not aligned to cues (third laboratory exploratory). Difference: uncued slow minus band. Identification: share of people whose own leave-one-run-out model predicted their held-out run best; chance is 1/$N$. Brackets: 95\% bootstrap CI over people.}
\end{table*}

\begin{figure*}[!t]
\centering
\includegraphics[width=0.82\linewidth]{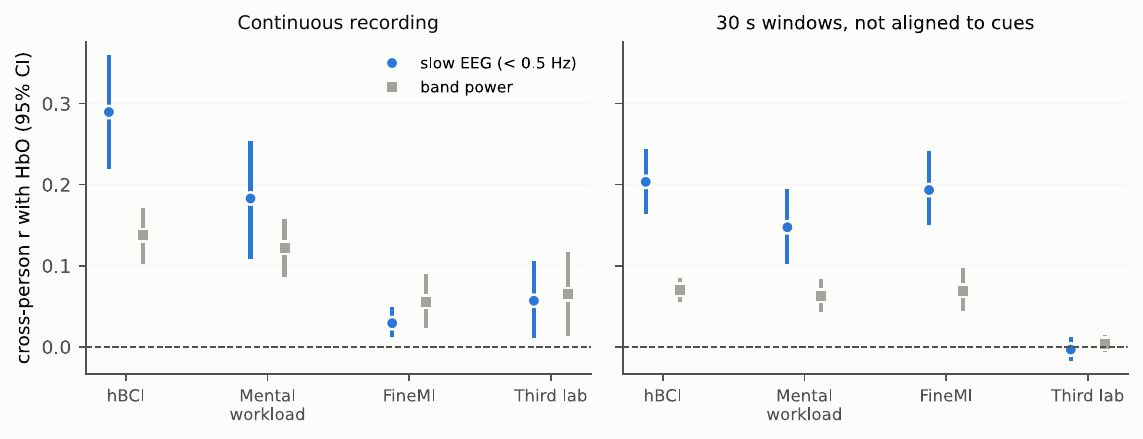}
\caption{Infra-slow EEG against band power under two methods. Left: the continuous method declared for the third laboratory. Right: 30~s windows not aligned to any cue. Error bars are 95\% CIs.}
\label{fig:method}
\end{figure*}

In hBCI, with a 0.5--45~Hz band-pass and 50~Hz notch, the deep pipeline predicted HbO across people at $r=0.019$; without the filter it gave $r=0.158$ [0.122, 0.191]. EEG below 0.5~Hz alone gave 0.155 (98\%), below 0.1~Hz 0.139, 0.1--0.5~Hz 0.053, and above 45~Hz 0.021. Ridge on infra-slow features outperformed the network (0.197; Table~\ref{tab:slow}).

Table~\ref{tab:slow} is best read column by column. In hBCI, MW and FineMI the slow-EEG value is at least twice the band-power value in both window types, and the uncued difference has a confidence interval well above zero in all three cohorts. Only the third laboratory breaks the pattern, with both feature sets at chance.

The advantage over band power replicated in MW and FineMI (Table~\ref{tab:slow}) and held on 30~s windows not aligned to cues (Fig.~\ref{fig:method}, right), although these still contain task periods. It depended on the analysis form: a continuous-recording method recovered it in hBCI but not in MW or FineMI (Section~\ref{sec:third}; Fig.~\ref{fig:method}, left).

Electrooculography channels alone carried 14\% of the hBCI score and $r=0.001$ in MW, and linear detrending raised the within-person score (0.360 to 0.404), so neither eye movements nor linear drift explain the effect. Much of the signal is global (hBCI: global 0.226, local 0.083), but a local component remained in all three cohorts ($r=0.07$--$0.16$).

Prediction was strongly person-specific. A person's own slow-EEG model scored 0.360 on their held-out run against 0.074 for other people's models in hBCI (difference $+0.286$ [0.234, 0.338]), 0.182 against 0.024 in MW and 0.252 against 0.093 in FineMI, and identified the person in 93\%, 62\% and 83\% of cases (Table~\ref{tab:slow}). Band power was far less person-specific (hBCI 0.074 against 0.014; 41\% identified).

\subsection{One EEG space for fNIRS and fMRI}
\label{sec:shared}

A rank-32 shared space predicted fNIRS in unseen people ($r=0.123$ in MW, $0.180$ in FineMI) and fMRI in 27 unseen people ($r=0.063$--$0.068$, $p=0.0001$) at no cost to either output: the 90\% CI lower bounds of the difference from modality-specific models were $-0.014$ and $-0.003$, above the $-0.02$ margin. It held for five of five seeds, on the local target and on uncued windows (fNIRS $r=0.134$ and $0.175$, fMRI $0.065$). The fMRI arm is the same in each test, and rank 32 was chosen after a declared rank-8 test failed on capacity. fMRI supervision did not improve fNIRS prediction over shuffled fMRI targets in any cohort.

\subsection{Calibrated monitoring and person-level reliability}
\label{sec:calibration}
\begin{table*}[!t]
\centering
\caption{Calibrated monitoring and reliability with two calibration runs.}
\label{tab:calib}
\footnotesize
\begin{tabularx}{\linewidth}{lXXXXX}
\toprule
Cohort & Gain, raw & Gain, local & $\rho$, cued & $\rho$, uncued & $\rho$, partial \\
\midrule
hBCI & \vc{+0.176}{0.135, 0.218} & \vc{+0.103}{0.074, 0.136} & \vc{0.68}{0.37, 0.86} & \vc{0.85}{0.66, 0.94} & \vc{0.84}{0.62, 0.94} \\
MW & \vc{+0.059}{0.025, 0.099} & \vc{+0.031}{0.013, 0.052} & \vc{0.56}{0.22, 0.78} & \vc{0.83}{0.64, 0.93} & \vc{0.83}{0.63, 0.94} \\
FineMI & \vc{+0.067}{0.033, 0.112} & \vc{+0.041}{0.014, 0.079} & \vc{0.90}{0.69, 0.97} & \vc{0.84}{0.57, 0.93} & \vc{0.69}{0.22, 0.89} \\
Third lab. & not run & not run & not run & \vc{0.29}{$-0.31$, 0.73} & not run \\
\bottomrule
\end{tabularx}
\tablenote{Gain: over replaying a person's own mean response. $\rho$: Spearman correlation between calibration score and later accuracy; partial: after removing HbO amplitude and EEG power. Brackets: 95\% CI.}
\end{table*}

\begin{figure*}[!t]
\centering
\includegraphics[width=0.82\linewidth]{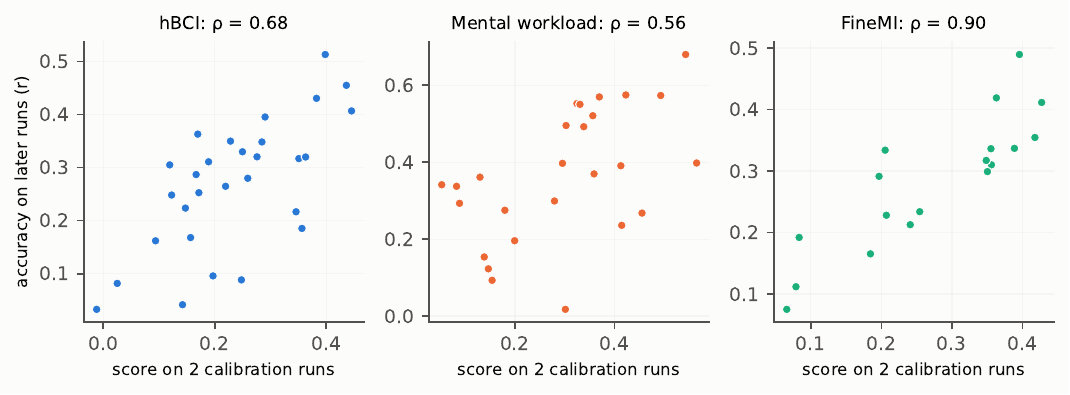}
\caption{Person-level reliability. Score on two calibration runs against accuracy on later runs, with one dot per person.}
\label{fig:calib}
\end{figure*}

After two calibration runs, the EEG-only model beat replaying the person's own mean response in all three cohorts, for raw and local targets (Table~\ref{tab:calib}); in hBCI the gain grew from $+0.118$ with one run to $+0.258$ with five.

The calibration score predicted later accuracy (Fig.~\ref{fig:calib}), surviving correction in all three cohorts and for five of five calibration-run seeds (four in MW), but not in the third laboratory (Table~\ref{tab:calib}). Trial-level reliability was not supported: the declared rule required hBCI to pass, and it did not ($-0.007$ [$-0.063$, 0.048]).

Rank correlations of 0.56--0.90 mean that people who were tracked well during calibration were mostly also tracked well later, and those near zero during calibration stayed low (Fig.~\ref{fig:calib}). The two calibration runs can therefore serve twice: to fit the personal model and to decide whether EEG-only monitoring is worth offering to that person. Figure~\ref{fig:signals}e shows a typical case. On the 20 trials of a later run, the personal model followed the measured channel-mean HbO with a mean per-trial $r$ of 0.55, whereas replaying the calibration mean response gave 0.09.

\subsection{Coupling and delay}
\label{sec:coupling}
\begin{table*}[!t]
\centering
\caption{Band-power coupling to local HbO.}
\label{tab:coupling}
\footnotesize
\begin{tabularx}{\linewidth}{lXXXX}
\toprule
Cohort & Beat own null & Median delay (s) & Personal delay repeats & Kernel repeats \\
\midrule
hBCI & 29/29 & +6 [3, 7] & Yes & Yes, $p<0.0001$ \\
MW & 81--85\% & +4 [1.5, 7.5] & No & Nominal, $p=0.006$ \\
FineMI & 83--89\% & +3 [$-1.5$, 5] & No (inconclusive) & Yes, $p=0.001$ \\
Third laboratory & 9/15 & +7 [2, 11] & No, $p=0.15$ & No, $p=0.10$ \\
\bottomrule
\end{tabularx}
\tablenote{Median delay with its 95\% CI; positive means EEG first. Ranges: raw and task-removed analyses. Kernel $p$ values are with the task response removed, except in the third laboratory.}
\end{table*}

\begin{figure*}[!t]
\centering
\includegraphics[width=0.55\linewidth]{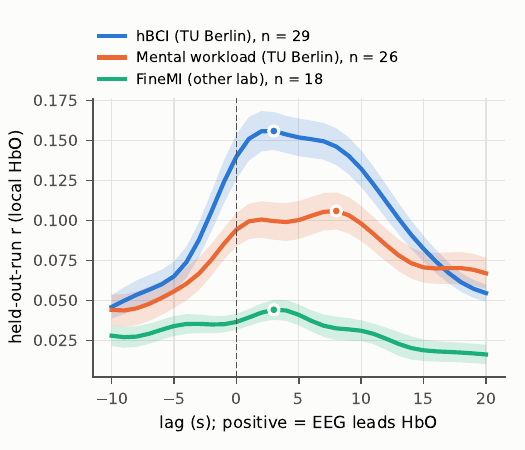}
\caption{Lag curves $R(\tau)$ from band power to local HbO (mean and standard error). Positive lag: EEG leads. Markers show group-mean peaks.}
\label{fig:lag}
\end{figure*}

\begin{figure*}[!t]
\centering
\includegraphics[width=0.82\linewidth]{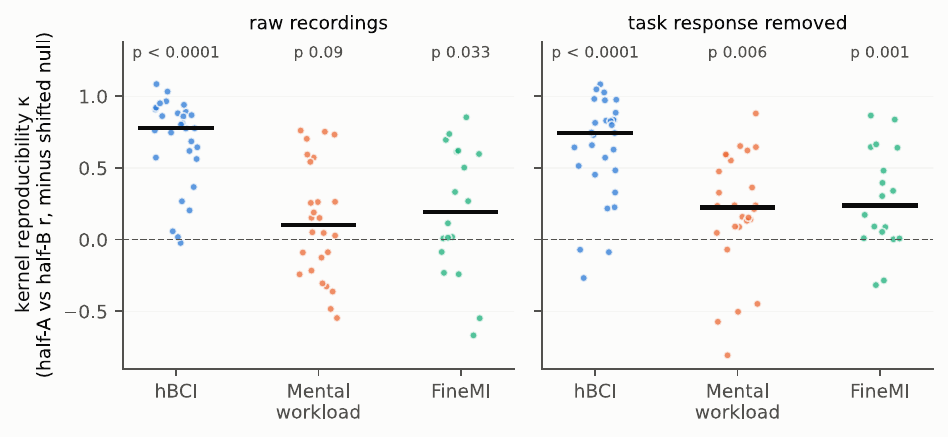}
\caption{Within-person kernel reproducibility, $\kappa$ of Eq.~(\ref{eq:kappa}), one dot per person. Left: raw recordings. Right: cue-locked response removed.}
\label{fig:kernel}
\end{figure*}

Band power predicted local HbO with EEG leading, and every hBCI person beat their surrogates (Fig.~\ref{fig:lag}, Table~\ref{tab:coupling}), with a median personal delay of $+6$~s (interquartile range 2--8~s; Wilcoxon $p=4\times10^{-6}$). Personal delays from the two halves of a recording agreed within 2~s in 66\% of people against 29\% by chance ($p=0.0003$), and in 69\% with the cue-locked response removed ($p=0.0001$), so the coupling is not only a shared task response.

The group-mean curves in Fig.~\ref{fig:lag} peak at positive lags in all three cohorts, at about 3~s in hBCI and FineMI and 8~s in MW, and fall off on both sides, as expected when a smoothed haemodynamic response follows electrical activity. Personal peaks are more spread out, which is why the median personal delay in Table~\ref{tab:coupling} differs from the peak of the mean curve. In Fig.~\ref{fig:kernel} most hBCI people lie well above zero in both panels, so their kernel shape repeated between the two halves of the recording; in MW and FineMI the dots scatter around a smaller median.

The kernel shape repeated beyond the task response in hBCI and FineMI, and nominally in MW (Fig.~\ref{fig:kernel}), but outside hBCI the peak delay did not repeat. Weaker coupling can explain this (median peak $r$ 0.19, 0.14 and 0.05 in hBCI, MW and FineMI): hBCI degraded with noise to FineMI's coupling strength lost its repeating delay (10\% against 16\% by chance), whereas hBCI cut to FineMI length kept it (66\% against 29\%). Infra-slow EEG itself appeared to follow HbO by 1--2~s in both TU Berlin cohorts, the reverse of the neurovascular order, but only nominally ($p=0.006$ and $0.004$ against $\alpha=0.0036$), and this was lost in MW under a later amendment.

\subsection{Transfer between cohorts}
\label{sec:transfer}
\begin{table*}[!t]
\centering
\caption{Transfer of models frozen on one cohort.}
\label{tab:transfer}
\footnotesize
\begin{tabularx}{\linewidth}{Xll}
\toprule
Test & $r$ [95\% CI] & Outcome \\
\midrule
hBCI to MW & +0.205 [0.128, 0.285] & Holds, survives \\
MW to hBCI & +0.177 [0.143, 0.211] & Holds, survives \\
hBCI to MW, without five closest pairs & +0.217 [0.119, 0.305] & Holds, survives \\
hBCI to MW, windows not aligned to cues & +0.201 [0.128, 0.276] & Holds, survives \\
hBCI to FineMI, TU Berlin filter emulated & $-0.064$ [$-0.094$, $-0.037$] & Fails, negative \\
hBCI to FineMI, matched window and montage & $-0.010$ [$-0.041$, +0.021] & Fails \\
hBCI to third laboratory, frozen kernel & $-0.026$ [$-0.059$, +0.009] & Fails \\
ds002336 to ds002338, fMRI output & +0.017 [$-0.009$, +0.042] & Fails \\
\bottomrule
\end{tabularx}
\end{table*}

\begin{figure*}[!t]
\centering
\includegraphics[width=0.55\linewidth]{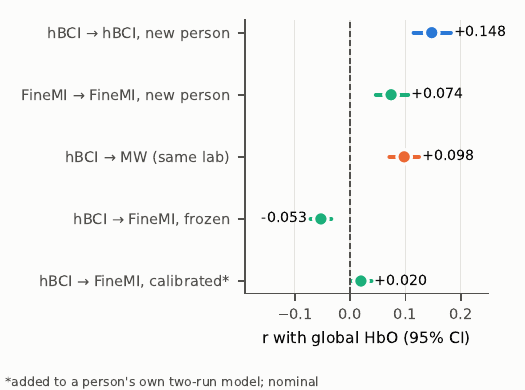}
\caption{Generalisation of the band-power coupling kernel within and across laboratories. Error bars: 95\% CIs. The calibrated gain (*) is the gain added to a person's own two-run model.}
\label{fig:transfer}
\end{figure*}

Models frozen on hBCI predicted MW, a different task in the same laboratory, and the reverse also held (Table~\ref{tab:transfer}). Adding the frozen model to a person's own two-run model raised MW accuracy from 0.319 to 0.378 (gain $+0.059$ [0.023, 0.095]) but added nothing in FineMI. EEG spectral fingerprints \cite{DaSilva2021} found five mutually best-matching cross-cohort pairs, about the number expected by chance, so shared participants were neither shown nor excluded; transfer held without them.

The frozen band-power kernel transferred within the laboratory (hBCI to MW, $+0.098$ [0.071, 0.124]) but was reliably negative in FineMI ($-0.053$ [$-0.071$, $-0.034$]; Fig.~\ref{fig:transfer}). Cross-laboratory transfer failed under every harmonisation tried: detrending, linked-mastoid referencing, Euclidean alignment, hardware-filter emulation, matched windows and montage, and a correlation-based signal improvement target. Within each laboratory, a kernel from other people did predict a new person (hBCI $r=0.148$, FineMI $0.074$). Removing overall EEG power only reduced the inversion ($r=-0.036$).

Figure~\ref{fig:transfer} summarises the pattern. Every transfer that stays within one laboratory, whether to a new person or to a different task, has a confidence interval above zero; the frozen transfer into the other laboratory falls below zero; and the calibrated gain in FineMI is close to zero. The coupling therefore generalises across people and tasks within a site, but not across sites.

\subsection{Decoding and the deep model}
\label{sec:downstream}

Predicted haemodynamics did not improve EEG-only classification beyond same-dimension principal-component or random-projection codes of the same infra-slow EEG (hBCI motor imagery 0.697 against 0.760 and 0.707, chance 0.50; likewise for mental arithmetic, $n$-back and FineMI). The benefit comes from the band itself (band power 0.61, infra-slow features 0.70--0.76). With the infra-slow band retained, fMRI supervision did not improve the deep network's fNIRS prediction ($-0.004$ [$-0.017$, $0.008$]), and it failed transfer, delay, decoding and uncertainty tests. It was run with one of five declared seeds and had about 3,200 training windows per fold for 11.9 million parameters, so the result is not evidence about deep models in general.

\subsection{Confirmation in a third laboratory}
\label{sec:third}

Of six tests declared in advance ($\alpha=0.0083$; 15 people), only band-power coupling passed (Table~\ref{tab:coupling}). Slow EEG predicted global HbO weakly with the declared continuous method ($r=0.057$ [0.014, 0.103], $p=0.013$), no better than band power ($0.065$); kernel and delay reproducibility and kernel transfer failed (Supplementary Table S6). A declared check showed that this method recovers the advantage in hBCI (+0.152 [0.086, 0.220]) but not in MW or FineMI (Fig.~\ref{fig:method}, left), so by the rule fixed in advance the test is uninformative here. In an exploratory analysis the window method found no prediction from either feature set (Table~\ref{tab:slow}). The advantage is not established in this laboratory.

\subsection{Attempts to strengthen the supported results}
\label{sec:strengthen}
\begin{figure*}[!t]
\centering
\includegraphics[width=0.82\linewidth]{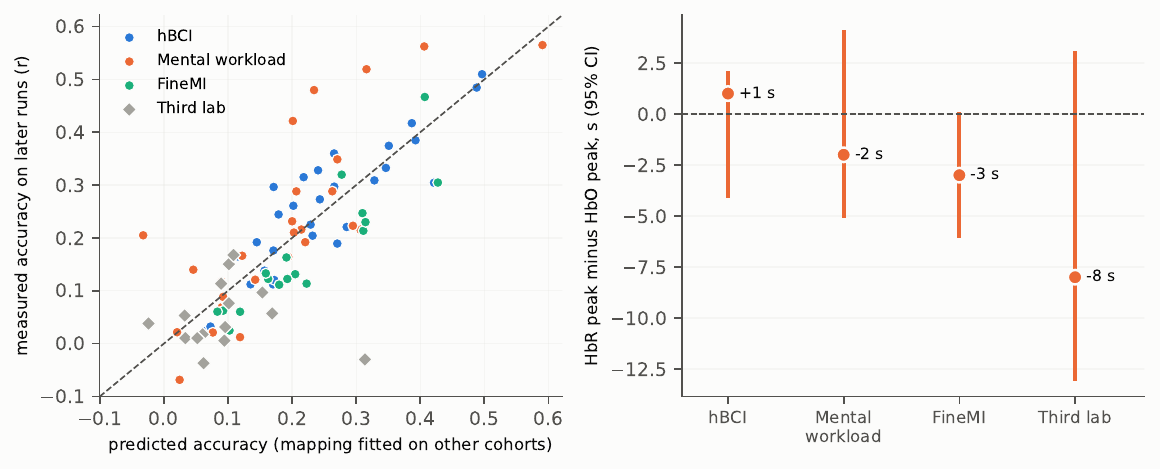}
\caption{Attempts to strengthen the supported results. Left: predicted against measured later accuracy in cohorts the map was not fitted on. Right: HbR peak minus HbO peak in four cohorts, with 95\% CIs.}
\label{fig:improve}
\end{figure*}

Fifteen further tests were declared ($\alpha=0.0033$); seven survived correction. Pooling both TU Berlin cohorts did not beat the better single-source model for either target (MW: 0.193 against 0.204). The expected later HbR peak was unresolved (CIs 6--16~s wide; Fig.~\ref{fig:improve}, right). The reliability score held after partialling out HbO amplitude and EEG power (Table~\ref{tab:calib}), and the map from calibration score to later accuracy beat a constant prediction in held-out cohorts (mean absolute error 0.046--0.077 against 0.106--0.172; Fig.~\ref{fig:improve}, left), surviving correction except in MW ($p=0.017$). In the third laboratory it was worse than that cohort's own mean accuracy (0.074 against 0.048; post hoc). In the three main cohorts, predicted accuracies were on average within 0.05--0.08 of the measured ones, so the calibration score predicts not only the ranking of people but also roughly the accuracy they will reach.

\section{Discussion}
\label{sec:discussion}

\subsection{Implication for EEG preprocessing}
\label{sec:impl}

In hBCI nearly all the information in the unfiltered deep-network score lay in a band that a standard 0.5--45~Hz filter removes, and in three cohorts infra-slow EEG predicted HbO better than band power. Filtered analyses may therefore wrongly conclude that EEG cannot predict haemodynamics. The finding does not apply to recordings high-pass filtered at acquisition: a 10~s amplifier time constant or unfiltered storage sufficed here, whereas the two excluded datasets had no power below 0.5~Hz. Filter choice should be reported and tested \cite{Widmann2015}.

A simple check follows from Fig.~\ref{fig:signals}c. Before choosing a filter, plotting each recording's spectrum down to the lowest resolvable frequency shows whether infra-slow power was recorded at all. Datasets shared only after high-pass filtering, such as the two we excluded, cannot be analysed in this way, so we encourage data sharers to keep an unfiltered or minimally filtered version.

\subsection{Relation to previous work}
\label{sec:prior}

Our band-power results agree with the weak correlations between spontaneous EEG power and HbO reported earlier \cite{Keles2016} and with coherence between EEG and NIRS near 0.1~Hz \cite{Nikulin2014}. Previous prediction of fNIRS from EEG was within person and used 0.1--100~Hz EEG \cite{Sirpal2021}; our cross-person values are lower than our within-person ones, which matches the strong person specificity we found. Cross-person prediction of fMRI from EEG has been reported with deep models \cite{LiuSajda2019,Li2024}. Our fMRI values are small, but they come from a linear model that shares its EEG space with fNIRS, and our question was whether one representation can serve both outputs, not how far fMRI accuracy alone can be pushed.

\subsection{Physiological origin}
\label{sec:timing}

Infra-slow EEG appeared to follow HbO by 1--2~s (nominally), whereas band power led local HbO by several seconds. If real, this would fit a vascular or non-neuronal source for the slow potential \cite{Voipio2003} alongside classical neurovascular coupling \cite{Iadecola2017}. Because much of the predictable signal is global, the slow relation may be substantially systemic \cite{Yuan2013,Tachtsidis2016}. We did not measure respiration, skin potential or scalp blood flow, and channel-mean removal is not equivalent to short-channel correction.

\subsection{Calibration and laboratory specificity}
\label{sec:route}

Cross-person and cross-laboratory models were weak or failed, whereas two calibration runs supported EEG-only monitoring, and the calibration score indicated beforehand who would be tracked well (except in the third laboratory). Calibration needs paired recordings from the same person, and accuracy remains modest (two-run personal model $r=0.32$ in MW, $0.10$ in FineMI). The kernel inversion between laboratories was broadband and only partly explained by global power; hardware, referencing, paradigm or systemic physiology may differ, so coupling models should be calibrated per site.

In practice the workflow is short. A session would begin with two runs recorded with both EEG and fNIRS; the personal model is fitted on them, the calibration score is computed, and EEG alone is used afterwards only for people whose score is high. No recordings from other people are needed at the point of use, although a population model, where available, can be added for the hybrid score.

\subsection{Limitations}
\label{sec:limits}

\begin{enumerate}
\item FineMI data quality. An internal audit before this cohort was adopted found a median scalp coupling index \cite{Pollonini2014} of 0.023, an evoked signal-to-noise ratio of 0.04 and split-half reliability near zero, and recommended exclusion. Its high slow-EEG prediction is therefore provisional, as poorly coupled optodes could share a slow non-cortical signal with EEG; without FineMI the advantage rests on the two TU Berlin cohorts.
\item Generality. The advantage is not established in the third laboratory and depended on the analysis form; the filtered-against-unfiltered test was run in hBCI with a deep network only, and the TU Berlin cohorts may share participants.
\item fMRI arm. All 27 people come from one study family, and fMRI prediction is small ($r$ about 0.065).
\item Analytic flexibility. Correction counts attempts rather than individual cohort tests, and several tests are amendments, including the rank-32 choice.
\item Seeds and causality. Most results use one seed, although ridge analyses are deterministic apart from bootstrap resampling. The analyses are predictive and do not establish a neural mechanism.
\end{enumerate}

\section{Conclusions}
\label{sec:conclusions}

In three cohorts from two laboratories, infra-slow EEG predicted HbO across people better than band power, and in hBCI almost all the information in an unfiltered deep-network score lay below 0.5~Hz, a band that conventional filtering removes. One EEG representation predicted both fNIRS and fMRI in unseen people at no cost to either output. Two calibration runs with paired recordings enabled EEG-only monitoring beyond a person's mean response and gave an advance estimate of its accuracy. Frozen models did not transfer between laboratories, and the advantage was not established in a third laboratory. Future work should measure systemic physiology directly and use short-separation fNIRS channels.

\section*{CRediT authorship contribution statement}
\textbf{Rafsan Jany:} Conceptualization, Methodology ,Supervision, Writing. \textbf{Muntasir Hossain:} Implementation, Validation, Formal analysis, Data curation, Writing. \textbf{Maisha Rahman:} Implementation,Validation, Visualization, Writing.

\section*{Declaration of competing interest}
The authors declare that they have no known competing financial interests or personal relationships that could have appeared to influence the work reported in this paper.

\section*{Declaration of generative AI and AI-assisted technologies in the manuscript preparation process}
During the preparation of this work, the authors used a generative AI assistant to assist with figure coding and manuscript drafting, as well as a generative AI image-generation tool to help design the graphical abstract. Following the use of these tools, the authors carefully reviewed and edited the generated content as necessary and take full responsibility for the final published article. All implementation, analyses, results, and conclusions are entirely the authors' own work and were conducted independently of generative AI.
\section*{Ethical approval}
This secondary analysis used public, de-identified data; the original studies obtained ethical approval and consent \cite{Shin2017,Shin2018,Yi2025,Chaiyanan2025,Lioi2020}.

\section*{Funding}
This research did not receive any specific grant from funding agencies in the public, commercial, or not-for-profit sectors.

\section*{Data availability}
All analysed recordings are public and cited in Table~\ref{tab:data}: hBCI \cite{Shin2017} and MW \cite{Shin2018} (TU Berlin), FineMI \cite{Yi2025}, learning capability \cite{Chaiyanan2025} (Mendeley Data, doi:10.17632/tsfs9fhn5y.1) and the EEG--fMRI studies \cite{Lioi2020} (OpenNeuro ds002336 and ds002338). Analysis scripts, the dated analysis plan, the multiplicity ledger and result files are available at \url{https://github.com/muntasirhossain2003/Cross-Modal-Prediction-of-fNIRS-Haemodynamics-From-Infra-Slow-EEG}.

\bibliographystyle{cas-model2-names}
\bibliography{references}

\end{document}